\documentclass[journal]{IEEEtran}

\usepackage{amsmath,amssymb,amsfonts}
\usepackage{amsthm}
\usepackage{algorithmic}
\usepackage{booktabs}
\usepackage{cite}
\usepackage{url}
\usepackage{graphicx}
\usepackage{microtype}

\newtheorem{theorem}{Theorem}
\newtheorem{proposition}{Proposition}
\newtheorem{corollary}{Corollary}

\title{Information-Geometric First-Passage Monitoring of Distributional Stability in Stochastic Systems}

\author{Hikmat Karimov and Rahid Zahid Alekberli%
\thanks{H. Karimov and R. Z. Alekberli are with the Institute of Defense Technologies and Cybersecurity, Azerbaijan Technical University (AZTU), Baku, Azerbaijan. Corresponding author: H. Karimov (e-mail: hikmat.karimov@aztu.edu.az).}%
\thanks{A preliminary conceptual formulation of this research was posted as arXiv:2604.24083. The present manuscript substantially revises the formal claims, sequential calibration, operational architecture, and empirical validation.}}

\begin{document}
\maketitle

\begin{center}
\footnotesize\emph{This work has been submitted to the IEEE for possible publication. Copyright may be transferred without notice, after which this version may no longer be accessible.}
\end{center}
\vspace{0.25em}

\begin{abstract}
Runtime monitoring of stochastic systems must distinguish nominal distributional relaxation from regime departure while controlling repeated-test false alarms under explicit validity assumptions. This paper links relative-entropy dissipation, information geometry, and sequential inference in a bounded first-passage monitoring architecture. For reversible Fokker--Planck dynamics, relative entropy to an invariant density is non-increasing; under exogenous forcing, its derivative decomposes into nominal dissipation and an information-space forcing term. The runtime layer uses Gaussian window surrogates, nominal-relative covariance shrinkage, a coordinate-consistent relative precision diagnostic, and randomized conformal ranks aggregated by a mixture power-martingale process. Analytical Ornstein--Uhlenbeck validation gives zero positive nominal Kullback--Leibler increments, forcing-identity residuals below $3.31\times10^{-6}$, and coordinate-invariance errors at numerical roundoff. On NSL-KDD, the monitor yields 0/100 alarms on internal nominal streams but 63/100 on official test-normal streams; post-change detection is 99.0\% for seen and 98.53\% for test-only attack types with median one-window delay. On UNSW-NB15, internal-null alarms are 0/100, whereas official test-normal alarms rise to 90/100; post-change detection is 81.33\%, with 18.67\% pre-change alarms. In these evaluations, calibration transport emerges as a major deployment constraint. No universal benchmark superiority, causal inference, or physical-work interpretation is claimed.
\end{abstract}


\section{Introduction}
\label{sec:intro}

Runtime monitoring in autonomous and cyber-physical systems increasingly operates on stochastic, high-dimensional telemetry rather than on deterministic state trajectories alone. In such settings, a useful monitor must distinguish ordinary distributional relaxation from sustained regime departure, remain interpretable under changes of representation, and control repeated-test false alarms under explicit null assumptions. These requirements are individually well studied across stochastic dynamics, information geometry, anomaly detection, and sequential inference. The present work addresses their joint treatment within one operational construction.

Relative entropy provides a natural distribution-level discrepancy, while reversible Fokker--Planck dynamics provide conditions under which relative entropy to an invariant density decreases along nominal evolution \cite{karatzas2022trajectorial,ito2024geometric}. Fisher information supplies a local geometric description of statistical sensitivity \cite{ay2017information,nielsen2020elementary}. Recent monitoring work has used geometric and distributional scores for anomaly or out-of-distribution detection \cite{legaria2023anomaly,fan2024simdetector,guo2025dcc,sun2026uav}, while sequential inference has developed time-uniform tools for repeated decisions \cite{vovk2021testing,howard2021timeuniform,ramdas2023savi,prinster2025watch}. Thus neither Kullback--Leibler (KL) divergence, Fisher geometry, covariance shrinkage, conformal martingales, nor first-passage testing is new in isolation.

This paper develops the Kerimov--Alekberli first-passage (KA-FPT) framework as a runtime architecture connecting these ingredients under explicit assumptions. The theoretical object is
\begin{equation}
D_t=D_{\mathrm{KL}}(p_t\Vert\pi),
\end{equation}
the relative entropy between the current state density and a nominal invariant density. Under reversible nominal dynamics, its derivative is a nonpositive dissipation functional. Under an added exogenous vector field, it decomposes into nominal dissipation and an information-space forcing contribution. This identifies a precise condition for divergence growth without interpreting that growth as physical work, malicious intent, or causality.

The runtime implementation is intentionally separated from the continuous-time theorem. Windowed features are represented by local Gaussian surrogates stabilized by nominal-relative covariance shrinkage, and a relative precision index is retained only as an auxiliary coordinate-consistent sensitivity statistic. Reference fitting, calibration, and online monitoring are disjoint. For finite-data operation, randomized conformal ranks are accumulated by a mixture power-martingale e-process; its time-uniform false-alarm semantics apply only under the stated conformal-martingale null assumptions.

The contributions are fourfold:
\begin{enumerate}
    \item a monitoring-oriented formalization of relative-entropy dissipation together with an explicit dissipation--forcing decomposition for reversible Fokker--Planck dynamics under stated regularity and boundary conditions;
    \item a Fisher-motivated runtime sensitivity layer with nominal-relative covariance shrinkage and a relative precision index invariant under invertible linear feature transformations;
    \item a leakage-separated reference/calibration/monitoring architecture coupled to an anytime first-passage e-process with explicit validity boundaries; and
    \item a three-tier empirical evaluation that separates analytical identity verification, within-calibration null behavior, cross-split calibration transport, and attack sensitivity on NSL-KDD and UNSW-NB15.
\end{enumerate}

The experiments reveal an important deployment boundary. KA-FPT produces no alarms in 100 held-out nominal streams drawn from the corresponding training populations, but the official test-normal populations trigger frequently in both network benchmarks. Consequently, the empirical contribution is not a universal ``1\% false-alarm'' or best-detector claim; it is a demonstration that first-passage sensitivity and calibration transport must be evaluated separately.

The framework is deliberately narrower than the earlier conceptual formulation of the Kerimov--Alekberli model. It does not claim a universal thermodynamic law for digital systems, does not infer ethical or malicious intent, and does not equate distributional stability with Lyapunov stability of individual trajectories. A thermodynamic interpretation is retained only for physical microstate distributions satisfying the stated canonical assumptions.

\section{Related Work and Foundational Boundaries}
\label{sec:related}

\subsection{Information Geometry and Distributional Monitoring}
Information geometry equips parametric probability families with the Fisher information metric and divergence-induced local geometry \cite{ay2017information,nielsen2020elementary}. Recent work has already used non-Euclidean probability-space geometry for anomaly detection \cite{legaria2023anomaly}, dynamic covariance structure for out-of-distribution detection \cite{guo2025dcc}, and information-geometric constructions for UAV-swarm anomaly analysis \cite{sun2026uav}. Accordingly, Fisher geometry, covariance-based monitoring, and information-geometric anomaly detection are not claimed as new here. In KA-FPT, Fisher geometry supplies a representation-consistent local sensitivity layer accompanying the primary divergence monitor.

\subsection{Relative Entropy and Stochastic Dynamics}
Relative-entropy dissipation along diffusion dynamics is closely connected to gradient-flow descriptions of Fokker--Planck equations. Karatzas et al. \cite{karatzas2022trajectorial} relate relative entropy and relative Fisher information for Langevin--Smoluchowski diffusions. Ito \cite{ito2024geometric,ito2024correction} develops geometric thermodynamic links among Fokker--Planck dynamics, information geometry, optimal transport, and entropy production. Recent work further unifies divergences, generalized free energies, and Fokker--Planck structures \cite{lucchi2026unified}. Relative entropy is also established as a distributional-uncertainty device in stochastic control \cite{falconi2025lqg}. The present work therefore does not claim novelty in any generic KL--diffusion or KL--control connection. Its theoretical contribution is the explicit runtime decomposition of divergence change into nominal dissipation and exogenous forcing under the assumptions of Section~\ref{sec:theory}.

\subsection{KL-Based Anomaly Detection and Sequential Monitoring}
KL-based and geometric scores are established tools for online anomaly or shift detection \cite{fan2024simdetector,legaria2023anomaly}. Sequential and anytime-valid inference is likewise mature. Vovk \cite{vovk2021testing} develops online randomness testing with conformal martingales; Howard et al. \cite{howard2021timeuniform} give time-uniform confidence sequences; Ramdas et al. \cite{ramdas2023savi} formalize safe anytime-valid inference; and WATCH \cite{prinster2025watch} uses weighted conformal martingales for adaptive deployment monitoring. Related online change-detection approaches address industrial concept drift and distributed sensor networks \cite{zhang2025conformal,yang2024online}. KA-FPT therefore does not claim novelty in conformal martingales, alpha spending, or changepoint detection. Its contribution is their integration with the stochastic-dissipation and Fisher-relative monitoring layers under a strict calibration boundary.

\subsection{Covariance Estimation}
Covariance regularization under finite samples is a longstanding problem, with recent shrinkage estimators providing stable covariance and precision estimation in moderate and high dimensions \cite{ledoit2021shrinkage,ledoit2022quadratic}. The nominal-relative shrinkage used here is not proposed as a new general-purpose covariance estimator. Its role is structural: shrinking toward the nominal covariance preserves the covariance transformation law required by the relative precision diagnostic.

\subsection{Novelty Boundary}
Against this literature, the contribution of KA-FPT is the unified construction of four elements: (i) a nominal relative-entropy dissipation relation with explicit exogenous forcing decomposition; (ii) a coordinate-consistent Fisher-relative diagnostic supported by nominal-relative covariance regularization; (iii) a leakage-separated reference, calibration, and monitoring pipeline; and (iv) a sequential first-passage alarm whose false-alarm semantics are stated globally over the monitoring horizon. The framework should therefore be read as a structured runtime monitoring architecture rather than as a new divergence, a new Fisher metric, a new covariance estimator, or a new generic changepoint detector.

\section{Formal Theoretical Core: Information-Geometric First-Passage Stability Monitoring}
\label{sec:theory}

This section develops the formal core of the Kerimov--Alekberli (KA) monitoring framework. The monitored quantity is the relative entropy between the current state distribution and a nominal invariant reference. The framework separates nominal entropy dissipation, exogenous forcing, local Fisher geometry, and sequential first-passage calibration. None of the results below identifies physical causality or control-theoretic stability without the additional assumptions stated explicitly.

\subsection{Nominal Diffusion Model and Invariant Reference}
\label{subsec:nominal}

Let $\mathcal{X}\subseteq\mathbb{R}^{n}$ be the state space and let $p_t(x)$ denote a strictly positive probability density on $\mathcal{X}$. Let $\pi(x)>0$ be a normalized nominal invariant density. We impose the following assumptions.

\noindent\textbf{A1 (Regularity).} $p_t$ and $\pi$ are sufficiently smooth for the differentiations and integrations by parts used below, and $D_{\mathrm{KL}}(p_t\Vert\pi)<\infty$ over the monitoring interval.\par\smallskip

\noindent\textbf{A2 (Diffusion tensor).} $D(x)=D(x)^{\mathsf T}\succeq 0$ is measurable and sufficiently regular.\par\smallskip

\noindent\textbf{A3 (Reversible nominal dynamics).} The nominal forward dynamics admit the detailed-balance representation
\begin{equation}
\frac{\partial p_t}{\partial t}
=
\frac{1}{2}\nabla\!\cdot\!
\left(
p_t D \nabla \log\frac{p_t}{\pi}
\right).
\label{eq:reversible_fp}
\end{equation}

\noindent\textbf{A4 (Boundary condition).} Either $\mathcal{X}=\mathbb{R}^{n}$ with sufficiently fast decay at infinity, or the boundary satisfies the corresponding zero-flux condition so that boundary terms vanish.

Assumption A3 is a sufficient structural condition for the entropy-dissipation identity below. It deliberately excludes the claim that every Fokker--Planck system must admit this representation.

Define the nominal divergence score
\begin{equation}
D_t
\triangleq
D_{\mathrm{KL}}(p_t\Vert\pi)
=
\int_{\mathcal X}
p_t(x)\log\frac{p_t(x)}{\pi(x)}\,dx.
\label{eq:kl_monitor}
\end{equation}

\begin{theorem}[Nominal relative-entropy dissipation]
\label{thm:kl_dissipation}
Under Assumptions A1--A4,
\begin{equation}
\frac{d}{dt}D_{\mathrm{KL}}(p_t\Vert\pi)
=
-\mathcal I_D(p_t\Vert\pi)
\le 0,
\label{eq:kl_dissipation}
\end{equation}
where
\begin{equation}
\mathcal I_D(p_t\Vert\pi)
\triangleq
\frac{1}{2}
\int_{\mathcal X}
p_t
\left(
\nabla\log\frac{p_t}{\pi}
\right)^{\mathsf T}
D
\left(
\nabla\log\frac{p_t}{\pi}
\right)dx
\ge 0.
\label{eq:relative_fisher_dissipation}
\end{equation}
Hence $D_t$ is a non-increasing Lyapunov functional for the nominal distributional dynamics.
\end{theorem}

\begin{proof}
Since $\int_{\mathcal X} p_t\,dx=1$, differentiation of \eqref{eq:kl_monitor} gives
\begin{equation}
\frac{dD_t}{dt}
=
\int_{\mathcal X}
\frac{\partial p_t}{\partial t}
\log\frac{p_t}{\pi}\,dx.
\end{equation}
Substituting \eqref{eq:reversible_fp} and integrating by parts, with the boundary term eliminated by Assumption A4, yields
\begin{align}
\frac{dD_t}{dt}
&=
-\frac{1}{2}
\int_{\mathcal X}
p_t
\left(
\nabla\log\frac{p_t}{\pi}
\right)^{\mathsf T}
D
\left(
\nabla\log\frac{p_t}{\pi}
\right)dx \\
&=
-\mathcal I_D(p_t\Vert\pi).
\end{align}
Positive semidefiniteness of $D$ implies $\mathcal I_D\ge0$.
\end{proof}

Theorem~\ref{thm:kl_dissipation} is a statement about \emph{distributional relaxation relative to the specified invariant density}. It is not, by itself, a theorem of Lyapunov stability for an arbitrary physical state trajectory or closed-loop controller.

\subsection{Exogenous Forcing and Divergence-Growth Condition}
\label{subsec:forcing}

We next introduce an exogenous vector field $f_{\mathrm{ext}}(x,t)$ that perturbs the nominal density evolution:
\begin{equation}
\frac{\partial p_t}{\partial t}
=
\frac{1}{2}\nabla\!\cdot\!
\left(
p_t D \nabla\log\frac{p_t}{\pi}
\right)
-
\nabla\!\cdot\!\left(f_{\mathrm{ext}}p_t\right).
\label{eq:perturbed_fp}
\end{equation}
We additionally assume that the perturbation flux satisfies the same decay or zero-boundary-flux condition needed for integration by parts.

\begin{theorem}[Dissipation--forcing decomposition]
\label{thm:forcing_decomp}
Under Assumptions A1--A4 and \eqref{eq:perturbed_fp},
\begin{equation}
\frac{dD_t}{dt}
=
-\mathcal I_D(p_t\Vert\pi)
+
\mathcal R_t,
\label{eq:forcing_decomp}
\end{equation}
where
\begin{equation}
\mathcal R_t
\triangleq
\int_{\mathcal X}
p_t
\left(
\nabla\log\frac{p_t}{\pi}
\right)^{\mathsf T}
f_{\mathrm{ext}}(x,t)\,dx.
\label{eq:forcing_functional}
\end{equation}
\end{theorem}

\begin{proof}
The nominal term gives $-\mathcal I_D$ by Theorem~\ref{thm:kl_dissipation}. For the perturbation term,
\begin{align}
&-\int_{\mathcal X}
\nabla\!\cdot(f_{\mathrm{ext}}p_t)
\log\frac{p_t}{\pi}\,dx \\
&\qquad=
\int_{\mathcal X}
p_t
\left(
\nabla\log\frac{p_t}{\pi}
\right)^{\mathsf T}
f_{\mathrm{ext}}\,dx,
\end{align}
again by integration by parts and the assumed boundary condition.
\end{proof}

\begin{corollary}[Instantaneous divergence-growth condition]
\label{cor:divergence_growth}
Under the conditions of Theorem~\ref{thm:forcing_decomp},
\begin{equation}
\dot D_t>0
\quad\Longleftrightarrow\quad
\mathcal R_t>
\mathcal I_D(p_t\Vert\pi).
\label{eq:growth_condition}
\end{equation}
\end{corollary}

The quantity $\mathcal R_t$ is an \emph{information-space forcing contribution}. Calling it physical work would require an additional thermodynamic realization and is not assumed in the general monitoring model.

\subsection{Local Fisher Geometry and Representation Consistency}
\label{subsec:fisher}

Let $\{p_\theta:\theta\in\Theta\subseteq\mathbb{R}^{d}\}$ be a regular parametric statistical model with Fisher information matrix
\begin{equation}
G_{ij}(\theta)
=
\mathbb E_{p_\theta}
\left[
\frac{\partial\log p_\theta}{\partial\theta^i}
\frac{\partial\log p_\theta}{\partial\theta^j}
\right].
\label{eq:fim}
\end{equation}
For a local parameter displacement $\Delta\theta$,
\begin{equation}
D_{\mathrm{KL}}
\!\left(
p_\theta\Vert p_{\theta+\Delta\theta}
\right)
=
\frac{1}{2}
\Delta\theta^{\mathsf T}
G(\theta)
\Delta\theta
+
o(\|\Delta\theta\|^2).
\label{eq:kl_fisher_expansion}
\end{equation}

Accordingly, the Fisher--Rao line element is
\begin{equation}
ds_F^2
=
d\theta^{\mathsf T}
G(\theta)d\theta.
\label{eq:fisher_line}
\end{equation}
Because $G$ transforms as a metric tensor, $ds_F^2$ is invariant under smooth reparameterization of $\theta$. This is the representation-consistency statement used by the KA framework. No scalar invariance is attributed to $\det G$ or $\operatorname{tr}G$ in isolation.

For a multivariate Gaussian family $p=\mathcal N(\mu,\Sigma)$, the local Fisher--Rao line element is
\begin{equation}
ds_F^2
=
d\mu^{\mathsf T}\Sigma^{-1}d\mu
+
\frac{1}{2}
\operatorname{tr}
\left(
\Sigma^{-1}d\Sigma
\Sigma^{-1}d\Sigma
\right).
\label{eq:gaussian_fisher_metric}
\end{equation}

For runtime diagnostics we also define an auxiliary \emph{relative precision index}
\begin{equation}
S_F(k)
\triangleq
\frac{1}{d}
\operatorname{tr}
\left(
\Sigma_0\Sigma_k^{-1}
\right),
\label{eq:relative_precision_index}
\end{equation}
where $\Sigma_0$ is the nominal covariance. Under an invertible linear change of coordinates $y=Ax+c$, \eqref{eq:relative_precision_index} is preserved because $\Sigma_0'\Sigma_k'^{-1}=A(\Sigma_0\Sigma_k^{-1})A^{-1}$. Thus $S_F(k)$ is a coordinate-consistent scalar comparison relative to the nominal covariance. It is an auxiliary sensitivity summary, not a replacement for the full Fisher--Rao metric.

\subsection{Sequential First-Passage Boundary with Global False-Alarm Control}
\label{subsec:fpt}

A pointwise $(1-\alpha)$ threshold does not in general provide an $\alpha$-level guarantee for a monitor that is repeatedly evaluated over time. To obtain a genuine first-passage false-alarm guarantee, we allocate the total error budget across monitoring times.

Let $D_k$ denote the divergence score computed from monitoring window $k$ under nominal operation. Assume that for each $k$ there exist nominal quantities $\mu_k$ and $\sigma_k>0$ such that
\begin{equation}
\mathbb P_{\mathrm{nom}}
\left(
D_k-\mu_k\ge x
\right)
\le
\exp\left(
-\frac{x^2}{2\sigma_k^2}
\right),
\qquad x\ge0.
\label{eq:subgaussian_tail}
\end{equation}

Choose an error-spending sequence $\{\alpha_k\}_{k\ge1}$ satisfying
\begin{equation}
\alpha_k>0,
\qquad
\sum_{k=1}^{\infty}\alpha_k\le\alpha.
\label{eq:alpha_spending}
\end{equation}
Define the sequential boundary
\begin{equation}
b_k
\triangleq
\mu_k
+
\sigma_k
\sqrt{
2\log\frac{1}{\alpha_k}
},
\label{eq:sequential_boundary}
\end{equation}
and the first-passage alarm time
\begin{equation}
\tau_\alpha
\triangleq
\inf
\left\{
k\ge1:
D_k\ge b_k
\right\}.
\label{eq:fpt}
\end{equation}

\begin{theorem}[Time-uniform false-alarm control]
\label{thm:sequential_fa}
If \eqref{eq:subgaussian_tail} holds for every $k$ and \eqref{eq:alpha_spending} is satisfied, then
\begin{equation}
\mathbb P_{\mathrm{nom}}
\left(
\tau_\alpha<\infty
\right)
\le\alpha.
\label{eq:global_fa}
\end{equation}
No independence assumption between monitoring windows is required.
\end{theorem}

\begin{proof}
By \eqref{eq:subgaussian_tail} and \eqref{eq:sequential_boundary},
\begin{equation}
\mathbb P_{\mathrm{nom}}(D_k\ge b_k)
\le\alpha_k.
\end{equation}
Therefore, by the union bound,
\begin{align}
\mathbb P_{\mathrm{nom}}(\tau_\alpha<\infty)
&=
\mathbb P_{\mathrm{nom}}
\left(
\bigcup_{k\ge1}\{D_k\ge b_k\}
\right) \\
&\le
\sum_{k\ge1}
\mathbb P_{\mathrm{nom}}(D_k\ge b_k) \\
&\le
\sum_{k\ge1}\alpha_k
\le\alpha.
\end{align}
\end{proof}

For a fixed monitoring horizon $K$, the simple choice $\alpha_k=\alpha/K$ gives family-wise false-alarm control. For indefinite monitoring, one admissible spending rule is
\begin{equation}
\alpha_k
=
\frac{6\alpha}{\pi^2 k^2},
\label{eq:infinite_alpha_spending}
\end{equation}
because $\sum_{k\ge1}k^{-2}=\pi^2/6$.

Theorem~\ref{thm:sequential_fa} assumes that the nominal tail parameters are known or conservatively certified. When they are estimated from finite calibration data, estimation uncertainty must also be controlled. Split-conformal or anytime-valid conformal-martingale procedures provide alternative nonparametric routes, but their guarantees must be stated at the sequential rather than merely pointwise level.

\subsection{Conditional Thermodynamic Interpretation}
\label{subsec:thermo}

The general monitoring framework is probabilistic and does not require a thermodynamic interpretation. A physical free-energy interpretation is available only under additional assumptions.

Suppose $x$ represents a physical microstate of a system in contact with a heat bath at temperature $T$, with energy function $U(x)$ and canonical equilibrium density
\begin{equation}
\pi(x)
=
\frac{1}{Z}
\exp\left(
-\frac{U(x)}{k_BT}
\right).
\label{eq:canonical_pi}
\end{equation}
Define the nonequilibrium free-energy functional
\begin{equation}
\mathcal F[p]
=
\int_{\mathcal X}p(x)U(x)\,dx
+
k_BT
\int_{\mathcal X}p(x)\log p(x)\,dx.
\label{eq:noneq_free_energy}
\end{equation}

\begin{proposition}[Relative entropy as free-energy excess]
\label{prop:free_energy}
For every density $p$ with finite relative entropy to $\pi$,
\begin{equation}
\mathcal F[p]-\mathcal F[\pi]
=
k_BT
D_{\mathrm{KL}}(p\Vert\pi).
\label{eq:free_energy_kl}
\end{equation}
\end{proposition}

\begin{proof}
Using $\log\pi=-U/(k_BT)-\log Z$,
\begin{align}
k_BT D_{\mathrm{KL}}(p\Vert\pi)
&= k_BT\int p\log p\,dx + \int pU\,dx \\
&\quad + k_BT\log Z.
\end{align}
For the canonical equilibrium, $\mathcal F[\pi]=-k_BT\log Z$. Substitution yields \eqref{eq:free_energy_kl}.
\end{proof}

Equation~\eqref{eq:free_energy_kl} is not invoked for arbitrary learned embeddings, network-feature vectors, or abstract software states. Such variables acquire a thermodynamic interpretation only if an explicit physical realization connecting them to $U$, $T$, and the canonical reference measure is established.

\subsection{Interpretive Boundary}
\label{subsec:theory_boundary}

The theoretical core therefore links a nominal invariant density to relative-entropy dissipation, forcing-induced divergence growth, Fisher local sensitivity, and finally a sequential first-passage alarm.

Three distinctions are essential. First, distributional stability relative to $\pi$ is not identical to Lyapunov stability of a physical state trajectory. Second, the forcing contribution $\mathcal R_t$ is an information-space term and is not interpreted causally without an additional physical or structural model. Third, the thermodynamic identity in Proposition~\ref{prop:free_energy} applies only to physical realizations satisfying the canonical-equilibrium assumptions stated above.

\section{Operational Monitoring Architecture}
\label{sec:architecture}

The theoretical results in Section~\ref{sec:theory} concern probability densities. Runtime operation uses finite windows of projected telemetry and therefore requires an explicit estimation and calibration layer.

\subsection{Feature Projection and Local Gaussian Representation}
Let $x(t)\in\mathcal X\subseteq\mathbb R^n$ be raw telemetry and let $\phi:\mathcal X\rightarrow\mathbb R^d$ be a fixed feature map. At sample index $i$, set $z_i=\phi(x_i)$. With window length $W$ and stride $s$,
\begin{equation}
\mathcal Z_k=\{z_i:(k-1)s+1\le i\le(k-1)s+W\}.
\end{equation}
The local Gaussian approximation uses empirical mean $\widehat\mu_k$ and covariance $\widehat\Sigma_k$. It is an operational approximation, not an assertion that the underlying stream is globally Gaussian.

\subsection{Nominal-Relative Covariance Shrinkage}
To ensure positive definiteness without breaking linear coordinate equivariance, define
\begin{equation}
\widetilde\Sigma_k=(1-\lambda)\widehat\Sigma_k+\lambda\Sigma_0,
\qquad 0<\lambda<1,
\label{eq:shrinkage}
\end{equation}
where $\Sigma_0\succ0$ is the nominal reference covariance.

\begin{proposition}[Coordinate consistency of nominal-relative shrinkage]
\label{prop:shrinkage_invariance}
For any invertible affine feature transformation $z'=Az+c$, with $A\in\mathrm{GL}(d,\mathbb R)$,
\begin{equation}
\widetilde\Sigma_k'=A\widetilde\Sigma_kA^{\mathsf T}.
\end{equation}
Consequently,
\begin{equation}
S_F(k)=\frac{1}{d}\operatorname{tr}(\Sigma_0\widetilde\Sigma_k^{-1})
\label{eq:runtime_sf}
\end{equation}
is invariant under the transformation.
\end{proposition}

\begin{proof}
Because $\widehat\Sigma_k'=A\widehat\Sigma_kA^{\mathsf T}$ and $\Sigma_0'=A\Sigma_0A^{\mathsf T}$, Eq.~\eqref{eq:shrinkage} gives $\widetilde\Sigma_k'=A\widetilde\Sigma_kA^{\mathsf T}$. Thus $\Sigma_0'(\widetilde\Sigma_k')^{-1}=A(\Sigma_0\widetilde\Sigma_k^{-1})A^{-1}$, whose trace is unchanged.
\end{proof}

The runtime Gaussian divergence score is
\begin{align}
D_k=\frac{1}{2}\Big[&\operatorname{tr}(\Sigma_0^{-1}\widetilde\Sigma_k)
+(\widehat\mu_k-\mu_0)^{\mathsf T}\Sigma_0^{-1}(\widehat\mu_k-\mu_0)\nonumber\\
&-d+\log\frac{\det\Sigma_0}{\det\widetilde\Sigma_k}\Big].
\label{eq:runtime_kl}
\end{align}

\subsection{Disjoint Reference, Calibration, and Null-Control Sets}
Nominal training data are partitioned reproducibly as
\begin{equation}
\mathcal D_{\mathrm{nom}}
=\mathcal D_{\mathrm{ref}}\;\dot\cup\;\mathcal D_{\mathrm{cal}}\;\dot\cup\;\mathcal D_{\mathrm{null}}.
\end{equation}
Only $\mathcal D_{\mathrm{ref}}$ fits preprocessing, $\phi$, $\mu_0$, and $\Sigma_0$. Calibration windows from $\mathcal D_{\mathrm{cal}}$ initialize the conformal score sequence. $\mathcal D_{\mathrm{null}}$ is reserved for an internal null-control experiment and is never used to fit or calibrate the monitor.

\subsection{Randomized Conformal Ranks and Mixture E-Process}
The general error-spending theorem of Section~\ref{subsec:fpt} remains available when certified tail parameters are defensible. For finite-data experiments, however, deterministic empirical ranks had insufficient resolution for the planned $p$-series spending boundary. Before benchmark outcomes were inspected, the operational protocol was therefore frozen to a conformal-martingale implementation.

Let $C_1,\ldots,C_M$ be calibration-window scores and let $D_k$ be the current online score. Rank $D_k$ among the calibration scores and preceding online scores. Define
\begin{align}
A_k^{>} &= \sum_{m=1}^{M}\mathbf 1\{C_m>D_k\}
          +\sum_{i=1}^{k-1}\mathbf 1\{D_i>D_k\},\\
A_k^{=} &= \sum_{m=1}^{M}\mathbf 1\{C_m=D_k\}
          +\sum_{i=1}^{k-1}\mathbf 1\{D_i=D_k\}.
\end{align}
With independent $U_k\sim\mathrm{Unif}(0,1)$, the randomized conformal rank is
\begin{equation}
p_k=\frac{A_k^{>}+U_k(1+A_k^{=})}{M+k}.
\label{eq:randomized_conformal_p}
\end{equation}
For $\varepsilon\in\mathcal E=\{0.1,0.2,\ldots,0.9\}$ define the power martingale
\begin{equation}
M_k(\varepsilon)=\prod_{i=1}^{k}\varepsilon p_i^{\varepsilon-1}
\end{equation}
and the equal-weight mixture
\begin{equation}
E_k=\frac{1}{|\mathcal E|}\sum_{\varepsilon\in\mathcal E}M_k(\varepsilon).
\label{eq:mixture_eprocess}
\end{equation}
The first-passage alarm is
\begin{equation}
\tau_\alpha=\inf\{k:E_k\ge1/\alpha\}.
\label{eq:eprocess_stop}
\end{equation}
Under the stated exchangeability/conformal-martingale null assumptions and a fixed score function, the conformal-martingale construction follows the standard online-testing line represented by \cite{vovk2021testing,prinster2025watch}; Ville's inequality gives
\begin{equation}
\Pr_{\mathrm{null}}\!\left(\sup_k E_k\ge1/\alpha\right)\le\alpha.
\end{equation}
This guarantee is not transported automatically to a different nominal population.

\begin{figure}[t]
\footnotesize
\begin{algorithmic}[1]
\REQUIRE Frozen $\phi$, $(\mu_0,\Sigma_0)$, calibration scores, $W,s,\lambda,\alpha$
\STATE Initialize $\{M_0(\varepsilon)=1:\varepsilon\in\mathcal E\}$ and prior calibration scores.
\FOR{$k=1,2,\ldots$}
\STATE Form $\mathcal Z_k$ and compute $\widehat\mu_k,\widehat\Sigma_k$.
\STATE $\widetilde\Sigma_k\leftarrow(1-\lambda)\widehat\Sigma_k+\lambda\Sigma_0$.
\STATE Compute $D_k$ and $S_F(k)$ from Eqs.~\eqref{eq:runtime_kl}--\eqref{eq:runtime_sf}.
\STATE Compute randomized conformal $p_k$ using Eq.~\eqref{eq:randomized_conformal_p}.
\STATE Update $M_k(\varepsilon)$ and mixture $E_k$ in Eq.~\eqref{eq:mixture_eprocess}.
\IF{$E_k\ge1/\alpha$}
\STATE Return first-passage alarm $\tau_\alpha=k$.
\ENDIF
\STATE Append $D_k$ to the preceding score sequence and advance by stride $s$.
\ENDFOR
\end{algorithmic}
\caption{KA-FPT runtime monitor used in the finite-data experiments.}
\label{alg:kafpt}
\end{figure}

\section{Empirical Validation and Benchmark Experiments}
\label{sec:experiments}

The evaluation separates three questions: analytical correctness of the stochastic identities, within-population sequential null behavior, and sensitivity after transporting the frozen monitor to published benchmark test distributions. No network experiment is used to establish the thermodynamic interpretation of Section~\ref{sec:theory}.

\subsection{Frozen Execution Protocol}
The sequential monitoring rule was frozen as protocol version 2.3 before attack outcomes were evaluated, following the deterministic-rank feasibility audit. After preliminary execution, a reproducibility/fairness patch (version 2.3.1) separated stream-generation randomness from randomized-conformal tie breaking and enforced matched pseudo-streams across comparators; it did not alter the divergence score, windowing, covariance shrinkage, alarm threshold, or stream-construction rules. All results reported below were recomputed from scratch under version 2.3.1. For both network datasets, normal training records were divided 50\%/25\%/25\% into $\mathcal D_{\mathrm{ref}}$, $\mathcal D_{\mathrm{cal}}$, and $\mathcal D_{\mathrm{null}}$. Standardization and PCA were fitted only on $\mathcal D_{\mathrm{ref}}$; the retained dimension was the smallest $d$ with at least 92\% explained reference variance. We used $W=s=100$, $\lambda=0.10$, $\alpha=0.01$, 100 pseudo-streams per attack type/category, 10 pre-change windows, 20 post-change windows, and 30-window null streams. Test labels were used only by the offline stream generator and evaluator; the online monitor received unlabelled features.

A pre-execution feasibility audit showed that deterministic empirical-rank $p$-values combined with $p$-series alpha spending were unusable at the available calibration sizes. With the final calibration allocation, the minimum deterministic ranks were $1/169$ for NSL-KDD and $1/141$ for UNSW-NB15, whereas the first spending threshold at $\alpha=0.01$ is $6\alpha/\pi^2\approx0.00608$. The finite-data monitor was therefore frozen to the randomized conformal mixture e-process of Section~\ref{sec:architecture} before attack outcomes were computed.

\subsection{Controlled Ornstein--Uhlenbeck Validation}
The four-dimensional OU process used
\begin{equation}
B=\mathrm{diag}(0.5,0.8,1.0,1.2),\qquad L=I_4,
\end{equation}
with invariant covariance defined by $B\Sigma_0+\Sigma_0B^{\mathsf T}=LL^{\mathsf T}$. The nominal dissipation test started from a displaced Gaussian. For forcing tests, at $t^\star=5$ an additive direction $v=[1,1,1,1]^{\mathsf T}/2$ was applied with $\beta\in\{0.2,0.5,1.0,2.0\}$.

The nominal analytical trajectory had
\begin{equation}
E_{\mathrm{diss}}=\max_t[D_{t+\Delta t}-D_t]_+=0.
\end{equation}
Across the forcing sweep, the maximum absolute residual in $\widehat{\dot D}_t+\mathcal I_D-\mathcal R_t$ was at most $3.30\times10^{-6}$ and the largest RMSE was $5.57\times10^{-7}$. Final analytical divergence increased from approximately $0.05$ at $\beta=0.2$ to $4.96$ at $\beta=2.0$. Under three pre-specified invertible linear feature transformations, the largest absolute errors were $1.11\times10^{-16}$ for $S_F$ and $1.78\times10^{-15}$ for Gaussian KL.

A separate finite-ensemble study with 100 Monte Carlo runs and 1000 particles per run quantified plug-in estimator error rather than theorem error. Mean forcing-residual RMSE increased from $0.0697$ at $\beta=0.2$ to $0.2702$ at $\beta=2.0$.

\subsection{NSL-KDD}
The published NSL-KDD split contained 125,973 training and 22,544 test records, with 67,343 and 9,711 normal records, respectively. The frozen PCA rule retained $d=22$ components and explained 92.25\% of reference variance. Seen attacks comprised \texttt{neptune}, \texttt{smurf}, \texttt{satan}, and \texttt{ipsweep}; the unseen regime used 17 attack types present only in the test set.

On 100 internal null streams from $\mathcal D_{\mathrm{null}}$, KA-FPT produced no alarms. This empirical count does not itself verify a 1\% rate; the time-uniform guarantee remains conditional on the conformal-martingale null assumptions. When the same frozen monitor was transported to official test-normal records, 63/100 30-window streams alarmed; among alarmed streams the median first passage was window 26. Thus the published test-normal population is not treated as exchangeable with the training-normal calibration population.

For attack pseudo-streams, post-change detection was 396/400 (99.0\%) for seen attack types and 1675/1700 (98.53\%) for the 17 test-only attack types. Pre-change alarms occurred in 4/400 and 25/1700 streams, respectively, and the pooled median post-change delay was one window. Several rare test-only types had very few unique records and required sampling with replacement; their type-level estimates are therefore sensitivity demonstrations rather than precise population estimates.

\subsection{UNSW-NB15}
The standard UNSW-NB15 train/test partitions contain 175,341 and 82,332 records, including 56,000 and 37,000 normal records. The EVR rule retained $d=20$ components and explained 92.84\% of reference variance. Nine attack categories were evaluated: Exploits, Fuzzers, Generic, DoS, Reconnaissance, Analysis, Backdoor, Shellcode, and Worms \cite{moustafa2015unsw}.

Again, KA-FPT produced 0/100 alarms on internal null streams. In contrast, 90/100 official test-normal streams alarmed within 30 windows, with median first passage 16.5 among alarmed streams. Across 900 attack pseudo-streams, 732 (81.33\%) produced a post-change alarm and 168 (18.67\%) alarmed during the 10-window pre-change test-normal segment. Conditional on surviving the pre-change segment without an alarm, every remaining stream subsequently alarmed after attack injection; pooled median post-change delay was one window. Because the pre-change test-normal segment is itself shifted relative to training calibration, this demonstrates sensitivity rather than calibrated operational false-alarm control.

\begin{table}[t]
\caption{KA-FPT network evaluation summary.}
\label{tab:results}
\centering
\scriptsize
\begin{tabular}{@{}lrrrr@{}}
\toprule
Setting & Int. null & Test null & Post-change & Delay\\
\midrule
NSL seen & 0/100 & 63/100$^{a}$ & 396/400 & 1\\
NSL unseen & -- & -- & 1675/1700 & 1\\
UNSW categories & 0/100 & 90/100 & 732/900 & 1\\
\bottomrule
\multicolumn{5}{l}{\scriptsize $^{a}$Shared NSL external-null stress test; delay in windows.}
\end{tabular}
\end{table}

\begin{figure}[t]
\centering
\includegraphics[width=\columnwidth]{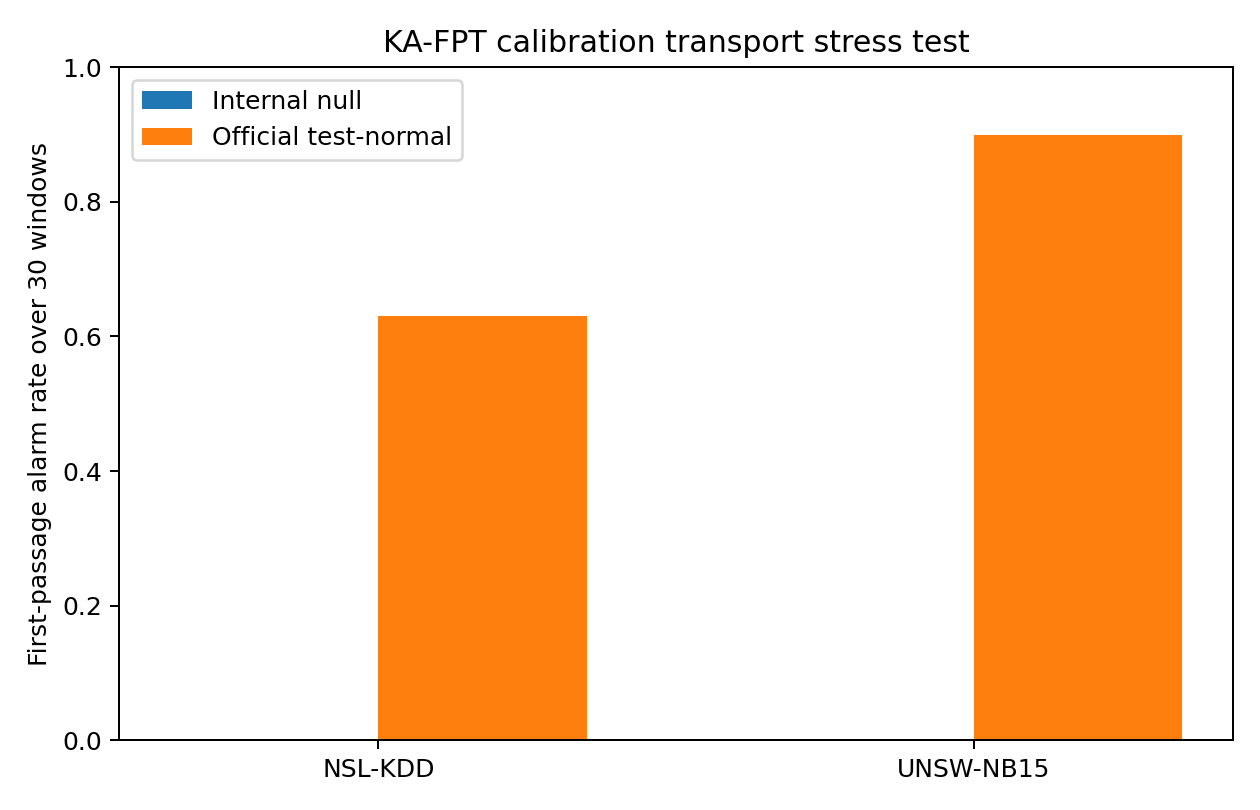}
\caption{First-passage alarm rate for held-out internal nominal streams versus official test-normal streams. The contrast reveals a calibration-transport failure relative to within-training-population null behavior.}
\label{fig:transport}
\end{figure}

\subsection{Comparator Interpretation}
The experiments do not establish uniform superiority over all baselines. On NSL-KDD, isolation forest produced 0\% external test-normal alarms and a mean attack-type detection probability of 85.7\%, compared with 63\% and 98.6\% for KA-FPT; its median detection delay was six windows versus one for KA-FPT. On UNSW-NB15, the covariance-entropy comparator produced 0\% external test-normal alarms and 89.4\% mean attack-category detection with median delay five windows, while KA-FPT produced 90\% external-null alarms, 81.3\% attack detection, and median delay one. The empirical contribution is therefore rapid distributional sensitivity with explicit first-passage semantics and an observable calibration-validity boundary, not a universal accuracy claim.

\section{Scope, Identifiability, and Limitations}
\label{sec:scope}

\subsection{Distributional Drift vs. Pointwise Stability}
The windowed score $D_k$ measures displacement of a fitted local distribution from a nominal reference. Small values do not imply Lyapunov stability, trajectory-wise robustness, or closed-loop asymptotic stability.

\subsection{Sensitivity to Nominal-Reference Estimation}
Gaussian KL depends nonlinearly on $(\mu_0,\Sigma_0)$ through quadratic, inverse-covariance, trace, and log-determinant terms. The dedicated reference/calibration split prevents direct optimistic reuse but does not eliminate finite-sample estimation uncertainty.

\subsection{Gaussian Projection and Shrinkage Boundary}
Let $p$ have finite mean $m_p$ and covariance $C_p$, let $g_{\mathrm{mom}}=\mathcal N(m_p,C_p)$, and suppose the divergences displayed below are finite. For Gaussian $q=\mathcal N(\mu_0,\Sigma_0)$,
\begin{equation}
D_{\mathrm{KL}}(p\Vert q)=D_{\mathrm{KL}}(p\Vert g_{\mathrm{mom}})+D_{\mathrm{KL}}(g_{\mathrm{mom}}\Vert q),
\end{equation}
so the exact moment-matched Gaussian term is a lower bound. The operational score instead uses the shrunk covariance $\widetilde\Sigma_k$; therefore no universal upper- or lower-bound relation between operational $D_k$ and the full non-Gaussian divergence is claimed. For multimodal or heavy-tailed windows, $D_k$ is a stable first- and second-moment monitoring statistic.

\subsection{Sequential Validity and Calibration Transport}
The e-process guarantee requires the conformal-martingale null assumptions for the score sequence. Non-overlapping windows remove shared observations but do not establish exchangeability. The large alarm rates on official test-normal streams in Fig.~\ref{fig:transport} empirically demonstrate this distinction: a guarantee calibrated for one nominal population is not inherited by a shifted population. Future deployment therefore requires explicit drift-aware recalibration or methods that justify validity under the relevant form of covariate shift.

\subsection{Benchmark Construction Limits}
NSL-KDD and UNSW-NB15 are benchmark datasets, not timestamped deployment traces. Detection delays are therefore reported in windows, not seconds. Rare NSL-KDD test-only attack types require repeated sampling for fixed-length pseudo-streams, limiting the precision of type-level estimates. UNSW-NB15 is used as a secondary, newer-than-NSL-KDD benchmark rather than as a representation of all current network threats.

\subsection{Physical and Interpretive Boundaries}
The free-energy identity $\mathcal F[p]-\mathcal F[\pi]=k_BT D_{\mathrm{KL}}(p\Vert\pi)$ is invoked only for physical microstate distributions satisfying the canonical isothermal assumptions of Section~\ref{sec:theory}. No physical-work interpretation is assigned to learned embeddings or packet features. The first-passage trigger does not identify physical cause, distinguish malicious intent from benign operating change, infer normative ethical properties, or authorize an intervention.

\section{Conclusion}
\label{sec:conclusion}

This work develops a bounded first-passage framework for monitoring distributional stability rather than a universal thermodynamic, control-stability, or ethical theory. The formal core establishes relative-entropy dissipation for a stated reversible Fokker--Planck class and decomposes divergence change under exogenous forcing into nominal dissipation and an information-space forcing contribution. Fisher geometry supplies a local representation-consistent sensitivity layer, while the operational monitor uses nominal-relative covariance shrinkage, a relative precision index, leakage-separated calibration, and a conformal-martingale first-passage e-process.

The analytical OU experiment supports the internal mathematical construction: nominal KL is monotone on the frozen grid, the forcing identity closes to residuals below $3.31\times10^{-6}$, and the proposed affine-coordinate diagnostics are preserved to numerical roundoff. The network benchmarks, however, expose a complementary limitation. KA-FPT produced 0/100 alarms on held-out nominal streams from the training populations, yet 63/100 NSL-KDD and 90/100 UNSW-NB15 official test-normal streams alarmed. Thus calibration transport across nominal populations is a first-order deployment problem, not a secondary implementation detail.

Attack sensitivity remained high on NSL-KDD and substantial on UNSW-NB15, generally with one-window median post-change delay, but these results are not interpreted as universal detector superiority or as empirical confirmation of a 1\% false-alarm rate on shifted test populations. Comparator behavior reinforces this conclusion: some baselines were markedly more robust to cross-split nominal shift, albeit often with slower detection.

The central outcome is therefore twofold. First, stochastic dissipation, Fisher-relative diagnostics, and anytime first-passage inference can be integrated into a mathematically bounded monitoring architecture. Second, the validity of the resulting alarm semantics is only as portable as the calibration assumptions. Future work should focus on drift-aware recalibration, covariate-shift-valid sequential inference, richer non-Gaussian score models, and evaluation on genuinely timestamped autonomous-system telemetry.

\section*{Acknowledgment}
OpenAI ChatGPT (GPT-5.6, accessed in September 2026) was used as an assistive research and manuscript-preparation tool for literature mapping, mathematical consistency checks, drafting and language refinement across the manuscript, generation and review of experimental Python code, and LaTeX assembly. The authors independently reviewed the resulting definitions, assumptions, proofs, citations, code, computations, interpretations, and final manuscript and take full responsibility for the submitted work.

\bibliographystyle{IEEEtran}
\bibliography{references}

\end{document}